\documentclass[10pt,twocolumn,letterpaper]{article}

\usepackage{wacv}
\usepackage{times}
\usepackage{epsfig}
\usepackage{graphicx}
\usepackage{amsmath}
\usepackage{amssymb}
\usepackage{booktabs}
\usepackage[utf8]{inputenc}
\usepackage{bbm}
\usepackage[pagebackref=true,breaklinks=true,colorlinks,bookmarks=false]{hyperref}

\usepackage{lmodern} 
\usepackage{amsmath}
\usepackage{amssymb}
\usepackage{graphicx,scalerel}

\usepackage{setspace}

\newcommand{\timestamp}{\tau}
\newcommand{\duration}{\delta}

\newcommand{\disp}{\gamma}
\newcommand{\durationsum}{\Delta}
\newcommand{\latentdim}{D}
\newcommand{\rigidtransfo}{\rho}

\def\wacvPaperID{1889} 

\wacvalgorithmstrack   

\wacvfinalcopy 

\ifwacvfinal
\usepackage[breaklinks=true,bookmarks=false]{hyperref}
\else
\usepackage[pagebackref=true,breaklinks=true,colorlinks,bookmarks=false]{hyperref}
\fi

\begin{document}

\title{Representing motion as a sequence of latent primitives, a flexible approach for human motion modelling}
\author{Mathieu Marsot$^1$~~~
Stefanie Wuhrer$^1$~~~
Jean-S\'{e}bastien Franco$^1$~~~
Anne-Hélène Olivier$^2$\\
$^1$ \small{Univ. Grenoble Alpes, Inria, CNRS, Grenoble INP\thanks{Institute of Engineering Univ. Grenoble Alpes}, LJK, 38000 Grenoble, France}\\
$^2$ \small{Univ. Rennes, Inria, CNRS, IRISA, M2S, 35000 Rennes, France}\\
{\tt\small firstname.lastname@inria.fr}
}

\maketitle
\thispagestyle{empty}








%


\begin{abstract}
   We propose a new representation of human body motion which encodes a full motion in a sequence of latent motion primitives. Recently, task generic motion priors have been introduced and propose a coherent representation of human motion based on a single latent code, with encouraging results for many tasks. Extending these methods to longer motion with various duration and framerate is all but straightforward as one latent code proves inefficient to encode longer term variability. Our hypothesis is that long motions are better represented as a succession of actions than in a single block. By leveraging a sequence-to-sequence architecture, we propose a model that simultaneously learns a temporal segmentation of motion and a prior on the motion segments. To provide flexibility with temporal resolution and motion duration, our representation is continuous in time and can be queried for any timestamp. We show experimentally that our method leads to a significant improvement over 
   state-of-the-art motion priors on a spatio-temporal completion task on sparse pointclouds. Code will be made available upon publication.

\end{abstract}

\section{Introduction}

3D human body motion modelling is an underlying issue in many computer vision and graphics problems like 3D pose estimation, 4D completion from sparse inputs, character animation or motion generation. These ill-posed problems often require prior knowledge on human motion to generate plausible solutions. Recently, several works have proposed to learn task generic priors of 3D human body motion by capturing information about pose changes over time \cite{xu2021exploring,li2021task,3DV118} which have shown great performances on many applications. We propose to improve on two aspect of these representations. First, we propose more flexibility in temporal resolution and motion duration, which makes our method easier to use as it does not need to be retrained from scratch for new inputs. Second, the reconstruction quality of existing priors tends to degrade quickly when considering longer motions. Combining temporal segments help to alleviate this degradation, but discontinuities in motion can appear due to motion averaging or normalization. To improve on this aspect, we learn a latent space on temporal segments jointly with segmentation parameters that allow for smooth transitions. This allows us to model a large variety of motion without restriction.  

Our hypothesis is that motion is better represented as a sequence of latent primitives than in a single one. Making an analogy between actions and latent primitives, in a dataset of $x$ different actions, a  motion prior needs $x^y$ different latent codes to represent all possible $y$ element sequences of actions while a sequential motion prior requires only $x$ latent codes. As $y$ increases with longer motions, the number of latent codes grows exponentially with motion duration for classical motion priors but remains constant in a sequential representation. 

Our model leverages a sequence-to-sequence (seq2seq) architecture that is flexible~\wrt the sequence length of the input and allows to directly encode the motion into a sequence of latent primitives. We then decode the sequence of latent primitives using a decoder implicit in time that outputs a parametric 3D human body model for any given time instant.

We show that this sequential prior has better generalization capacity than a baseline using a single latent code on a synthetic dataset. Additionally, we show that our method generalizes to motion duration outside the training set, and provides state-of-the-art results for a spatio-temporal completion task on sparse unordered point cloud from data acquired in a multi-view studio.

In summary, our main contributions are 
\begin{itemize}
    \item A novel motion representation using a sequence of latent primitives. 
    \item An implicit representation of the temporal dimension allowing for flexible temporal resolution. 
    \item An ablation and a comparative study showing significant improvement~\wrt existing motion priors. 
\end{itemize}

\section{Related Work}

Human motion has extensively been studied in disciplines including computer vision, animation, perception, and human-computer interaction with a variety of goals. We focus our review on works that propose motion priors for 3D human motion data densely captured in space and time.

First such models have focused on motion as a sequence of static poses. For instance, when generating dense human motion from sparse MoCap~\cite{Anguelov05, Loper14, mahmood2019amass, habermann2021} or from 2D video data~\cite{kanazawa19, zhang19}, these works use per frame input marker points or 2D images as input to reconstruct dense 4D motion data. In this way, a statistical human body model has been fitted to a corpus of motion, providing the community access to a large 4D dataset, which we leverage in our work~\cite{mahmood2019amass}.

Recently, there has been a surge of interest for data driven motion priors, with first approaches focusing on a few motion sequences or multiple sequences of the same actor~\cite{Akhter2012,boukhayma18,regateiro19,regateiro2021deep4d}. 

Following these works, there were attempts at generalizing to different body shapes and larger variety of motion. One line of work focuses on implicit spatio-temporal representations~\cite{niemeyer2019occupancy,rempe2020caspr}, with the advantage of being flexible~\wrt spatial and temporal sampling of the input data. These works have been specialized to human motions~\cite{jiang2021learning,jiang2022h4d}, but as they do not reduce the spatial dimensionality of the input, they are restricted to small temporal spans. 

Another line of work closer to ours uses spatially aligned data, where the input is aligned on a statistical human body model. These works focus mostly on skeletal motion, which has a low dimensionality, and allows to represent longer temporal spans. Some priors were presented in a task specific setting,~\eg 3D pose estimation from monocular video \cite{kocabas2020vibe,Zhang:ICCV:2021} or motion generation \cite{rempe2021humor,petrovich2021action,ghorbani2020probabilistic}. Two of these methods consider problems that are different from ours, but present ideas that inspired our work. Ghorbani~\etal~\cite{ghorbani2020probabilistic} synthesize motion variations by subdividing long motions into smaller segments. Subdividing motion into segments is an idea which we leverage in our work.
Petrovich~\etal~\cite{petrovich2021action} synthesize motions of a specific given action, and explicitly model the duration of a motion. We also explicitly model segment duration in our approach.

Finally some priors were presented as task generic~\cite{li2021task, xu2021exploring,3DV118} with encouraging results on different tasks with a single model. Closest to our work, these priors focus on encoding motion in a latent space and show that this representation can be leveraged in tasks such as 3D pose estimation or 4D motion completion. To encode motion, these methods either consider motion of fixed duration and framerate~\cite{li2021task,xu2021exploring} or temporally aligned motion using a dynamic time warping criterion~\cite{3DV118}. These priors can model up to $3s$ motion when considering global displacement~\cite{3DV118} and $4s$ motion without considering global displacement~\cite{xu2021exploring}, but they do not generalize easily to longer motion sequences. We show experimentally that our approach outperforms two of these methods for the task of motion completion~\cite{li2021task, xu2021exploring}; the third method~\cite{3DV118} is limited to motions performing a cyclic hip motion and cannot be applied in our scenario of various motion.

\section{Overview}

\begin{figure}
\includegraphics[width=\columnwidth]{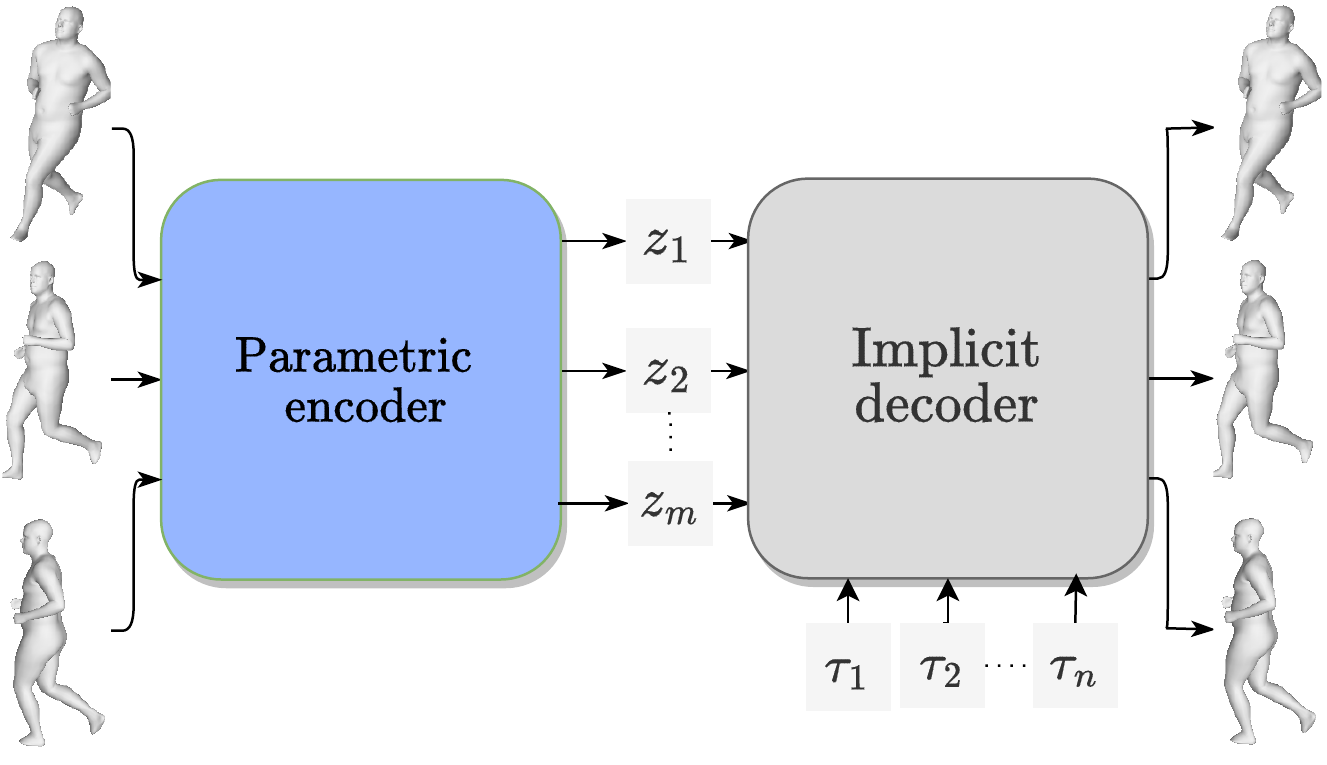}
\caption{Method overview. Architecture consists of a seq2seq encoder (blue) that maps a human motion sequence into a sequence of latent primitives $z_1,\ldots,z_m$, and a temporally implicit decoder (grey block) that decodes $z_1,\ldots,z_m$ and a series of timestamps $\timestamp_1,\ldots,\timestamp_n$ into a $n$-element sequence of parametric human body models.
}
\label{fig:archi}
\end{figure}

Figure~\ref{fig:archi} provides a visual overview of our method. We use a data driven approach to learn our motion representation. To deal with the high dimensionality of the spatio-temporal data used for training, we leverage a low-dimensional representation for each frame based on a human body model. 

Given as input a sequence of human motion with variable duration and number of frames, our goal is to encode the input directly in a sequence of latent primitives. To address this sequence to sequence problem, we choose the transformer architecture~\cite{vaswani2017attention}. This choice was guided by the transformer capability to better retain long term correlations than its recurrent network counterparts. 

To generate coherent output motions at arbitrary temporal resolution, we decode each latent primitive independently in a temporally implicit way, thereby converting the latent sequence to a sequence of parametric human body models for any given time instant. Decoding the primitives independently allows the model to build a single common latent space of motion.  We want the model to allocate each primitive to a temporal segment of the motion, without requiring handcrafted segmentation parameters, so the per-primitive outputs are combined using a weighted average computed using temporal masks. The temporal masks ensure that each primitive is allocated to a continuous temporal window of the input motion.

The model is trained as a conditional variational autoencoder (CVAE) using the shape conditioning proposed in prior works~\cite{3DV118} to retain correlations between body shape and motion.

\section{Method}

This section provides a formal description of the representation and architecture proposed to learn a sequence of latent primitives that represent human motion of variable duration, and provides details on the training.


\subsection{4D sequence representation}

We are interested in modelling a large variety of motions performed by different subjects. To do so, we leverage the AMASS dataset \cite{mahmood2019amass}. AMASS is an aggregation of multiple synthetic motion capture data for which a parametric body model is provided.


We consider body shape to remain constant over time, which allows to represent a motion sequence using a set of body shape parameters $\beta$, the joint rotations of a skeleton $\theta(\timestamp)$, and a 3D coordinate vector characterizing the displacement of the root joint $\disp(\timestamp)$, where $\timestamp$ is the parameter controlling time. For our implementation, we use the SMPL body model~\cite{loper2015smpl} provided with the dataset. More details are given in \ref{sec:implem}. 

We call $M(\theta(\timestamp),\disp(\timestamp),\beta)$ the function that outputs the template aligned mesh corresponding to the parametric representation $\theta(\timestamp),\disp(\timestamp),\beta$. A discretized motion sequence consisting of $n$ frames is then characterized by the sequence $\{M(\theta(\timestamp_i),\disp(\timestamp_i),\beta),\timestamp_i\}_{i=1}^n$, where $\timestamp_i$ are the time stamps corresponding to the meshes. 


\subsection{Transformer encoder}

\begin{figure}
    \centering
    \includegraphics[trim={1.7cm 0 0.5cm 0},clip,width=\columnwidth]{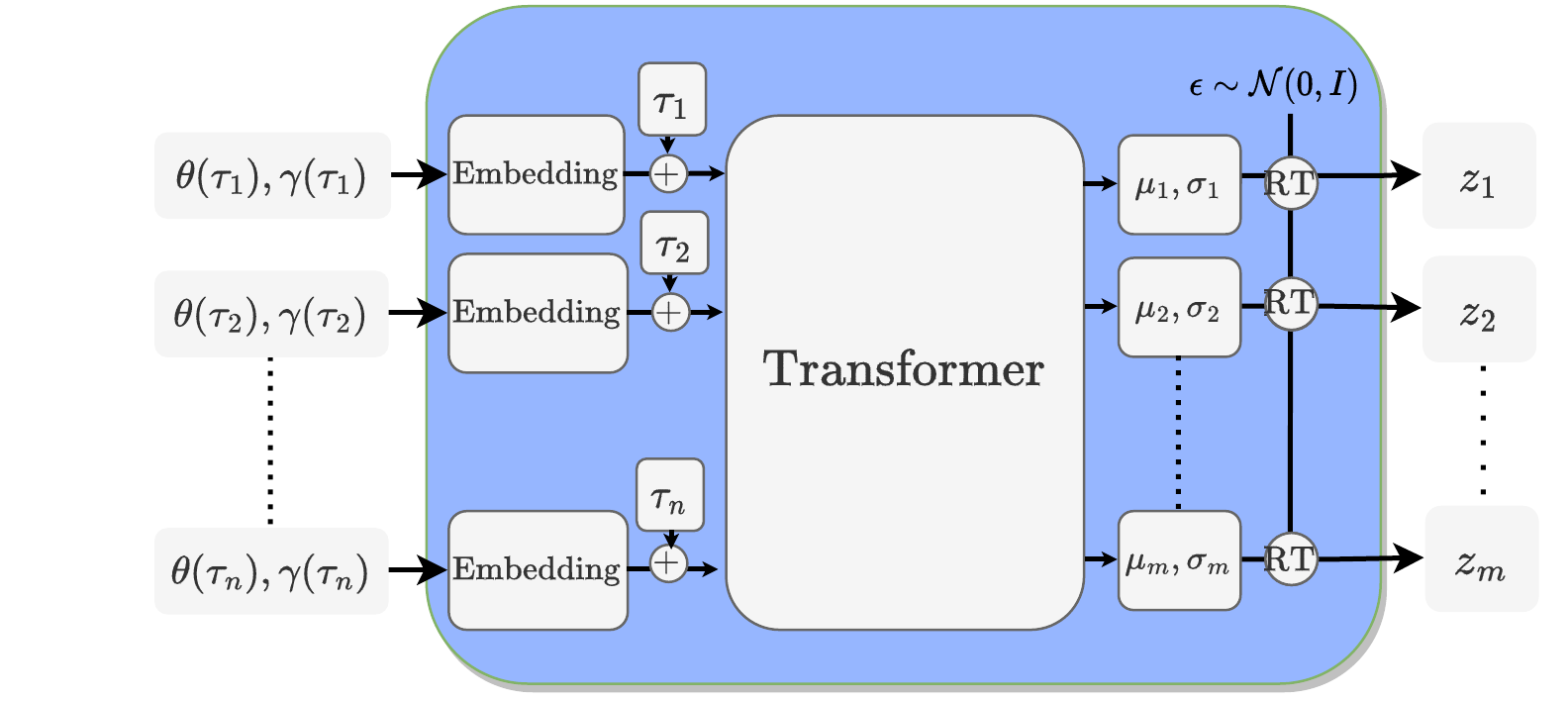}
    \caption{Encoder mapping an input sequence into a sequence of latent motion primitives $z_1,\ldots,z_m$. The embedding is a one layer perceptron. Time stamps $\timestamp_i$ are added as positional encoding. Transformer outputs a sequence latent distributions $\mu_i,\sigma_i$ from which $z_i$ are sampled using the reparametrization trick (RT).}
    \label{fig:encoder}
\end{figure}

We represent a 4D human motion sequence using two independent factors: a sequence of $m$ latent primitives $\{z_i\}_{i=0}^m$, and body shape $\beta$. To allow for flexible duration and frame rates, we allow an arbitrary number of input frames per sequence. The latent primitives are obtained by a transformer encoder that maps an input sequence to a sequence of latent primitives. The transformer block of the encoder operates on a reduced embedding representation of the input frames and is similar to the original transformer~\cite{vaswani2017attention} that was used for language translation. The encoder architecture is shown in Figure \ref{fig:encoder}. 
The difference between language translation and our setting is that we do not have ground truth for the latent primitives. To learn this part in an unsupervised setting, we fix the number $m$ of latent primitives but retain temporal flexibility by learning the duration $\duration_i$ of each motion primitive using a differentiable cost function.

The encoder considers a sequence of parameters $\{\theta(\timestamp_i),\disp(\timestamp_i),\timestamp_i\}_{i=1}^n$. To adjust the dimensionality of this representation, for each frame $\theta(\timestamp_i),\disp(\timestamp_i)$ are passed through one perceptron layer, and $\timestamp_i$ is subsequently concatenated to this representation. Note that the perceptrons applied to all frames share weights.

The output sequences of the transformer encoder are interpreted as a sequence of Gausssian distributions defined by a sequence of means and standard deviations $\{\mu_i,\sigma_i\}_{i=1}^m$ from which the latent primitives $\{z_i\}_{i=1}^m$ are sampled using Gaussian noise $\epsilon \sim \mathcal{N}(0,I)$ such that $z_i =\mu_i+\epsilon\sigma_i$. This is similar to the interpretation of latent spaces of VAEs and known to allow for generalization.


\subsection{Temporally implicit decoder}

\begin{figure}
    \centering
    \includegraphics[width=\columnwidth]{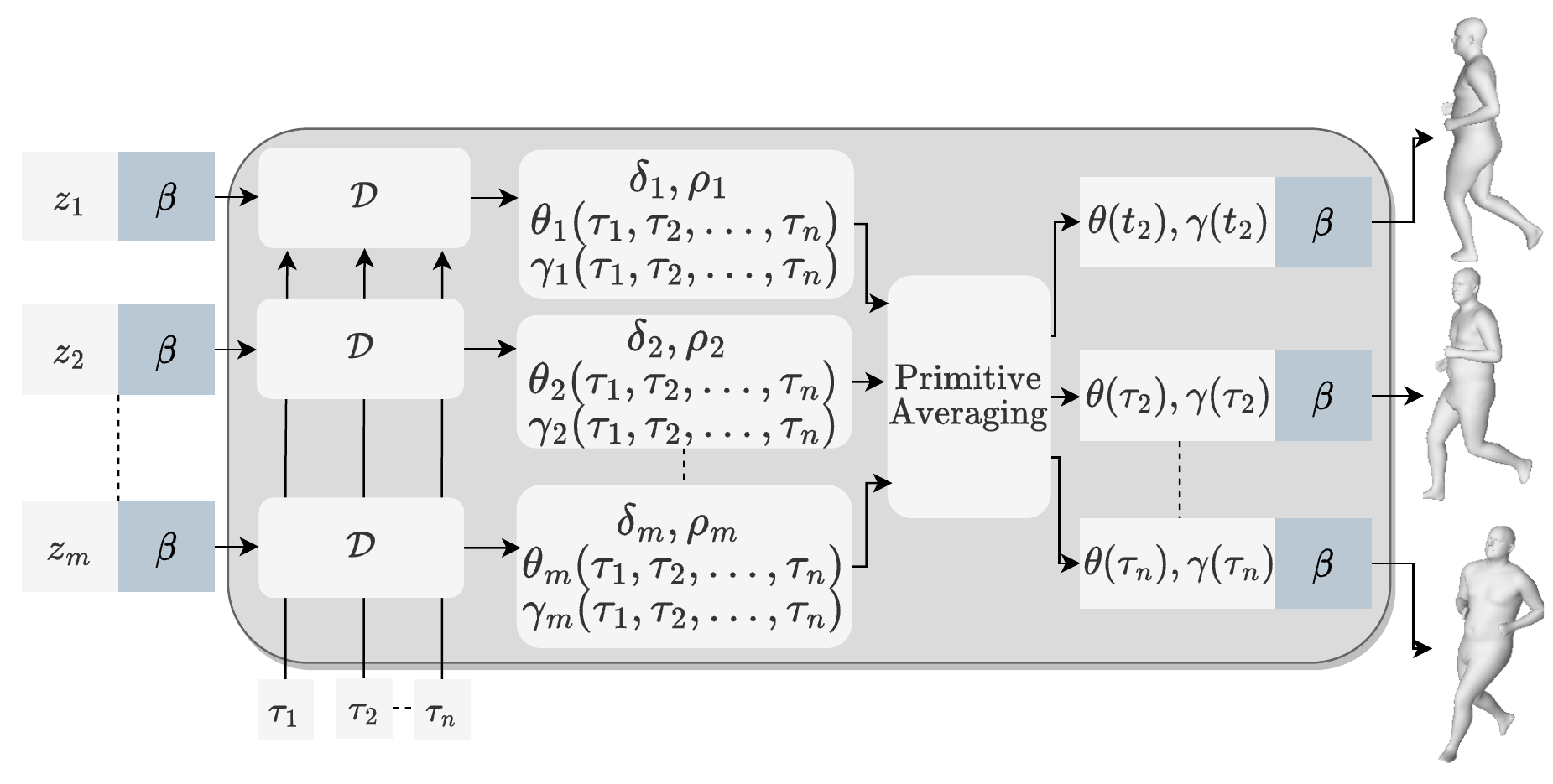}
    \caption{Implicit decoder. Given a sequence of $m$ latent primitives $z_i$, a body shape $\beta$ and a sequence of $n$ time\-stamps $\timestamp_j$, the decoder outputs a sequence of body meshes parameterized by $\beta,\theta,\disp$. The $z_i$ are decoded independently into segment parameters that include duration $\duration_i$ and a rigid transformation $\rigidtransfo_i$, and subsequently combined to decode a dense 4D motion. }
    \label{fig:decoder}
\end{figure}

Our decoder operates in two stages by first decoding individual latent primitives and by subsequently combining them. This allows learning a motion prior on the latent primitives which each characterize a motion segment, while still enforcing coherence between neighboring segments. The architecture of the implicit decoder is shown in Figure~\ref{fig:decoder}.

\textbf{Primitive decoding}
First, given a body shape $\beta$, the latent primitives are decoded individually using an implicit primitive decoder $\mathcal{D}(z_i,\beta,\timestamp)$ that outputs per-segment motion parameterized by $\theta_i(\timestamp),\disp_i(\timestamp)$, its duration $\duration_i$ and a rigid transformation $\rigidtransfo_i$. $\theta_i(\timestamp)$ and $\disp_i(\timestamp)$ characterize the global motion on the temporal segment [$\durationsum_i,\durationsum_i+\duration_i$], where $\durationsum_i = \sum_{j<i} \duration_j$. This ensures that each latent primitive encodes information of a continuous temporal segment of the input motion. For invariance~\wrt the initial orientation and displacement of a segment, $\mathcal{D}$ also outputs a rigid transformation $\rigidtransfo_i$ which is used as a transition from segment space to input space.
The architecture of $\mathcal{D}$ consists of two MLPs. The first MLP outputs per segment parametric representation $\theta_i(\timestamp)$,$\disp_i(\timestamp)$. The second MLP does not consider time and outputs the per segment parameters $\rigidtransfo_i$ and $\duration_i$.

\textbf{Primitive combination}
To combine the segment representations into motion $\theta(\timestamp),\disp(\timestamp)$, a weighted average of the per-segment representations using temporal masks is computed. For each primitive, the corresponding Gaussian mask is $G_i(\timestamp) = e^{-\big{(}\frac{\timestamp-\frac{(\durationsum_i+\duration_i)}{2})}{\duration_i/2}\big{)}^2}$ such that: 
\begin{eqnarray}
    \theta(\timestamp)& = &\frac{\sum_i G_i(\timestamp) (\rigidtransfo_i *\theta_i(\timestamp))}{\sum_i G_i(\timestamp)}, \\
    \disp(\timestamp)& = &\frac{\sum_i G_i(\timestamp) (\rigidtransfo_i *\disp_i(\timestamp))}{\sum_i G_i(\timestamp)}.
\label{eq:primtive_average}
\end{eqnarray}

We denote by $\rigidtransfo_i * \theta_i(\timestamp)$ and $\rigidtransfo_i *  \disp_i(\timestamp)$ the operation of applying the rigid transformation $\rigidtransfo_i$ to the corresponding body model parameters. This transformation consists of rotating the root joint for parameters $\theta$, and rotating and translating the global displacements $\disp$.

The temporally implicit nature of $\mathcal{D}$ alleviates the problem of averaging segments that may not be temporally aligned according to a predefined frame rate. The averaging of rotations is done in 6D representation space~\cite{zhou2019continuity}. This qualitatively leads to naturally combined results.

\subsection{Training}



The model is trained in a variational autoencoder setting with a reconstruction loss, and a Kullback–Leibler (KL) divergence loss to constrain the prior distribution to a normal distribution. We also added a regularization loss on the segment duration for faster convergence and to prevent local minima. 
The total loss is 
\begin{equation}
\mathcal{L} = \mathcal{L}_{rec}+\lambda_{KL}\mathcal{L}_{KL}+\lambda_{reg}\mathcal{L}_{reg},
\end{equation}
where 
\begin{eqnarray}
    \mathcal{L}_{rec} &=& \mathcal{L}_{global}+\mathcal{L}_{segment},\\
    \mathcal{L}_{KL} &=& \frac{1}{m}\sum_{j=1}^m KL(\mathcal{N}(\mu_j,\sigma_j),\mathcal{N}(0,I)),\\
    \mathcal{L}_{reg} &=&\sum_{j=1}^m (\duration_j-1/m)^2.
\end{eqnarray}
The reconstruction loss is divided into two terms. The first term $\mathcal{L}_{global}$ acts as a global reconstruction term between the input and the reconstructed output, including both a per vertex distance to capture fine details and a distance in the parametric representation. 
\begin{eqnarray}
    \mathcal{L}_{global}&=&\frac{1}{n}\sum_{i=1}^n \bigg{(}\Big{|}\Big{|}\theta(\timestamp_i)\hspace{-1mm}-\hspace{-1mm}\theta_{GT}(\timestamp_i)\Big{|}\Big{|}^2\hspace{-2mm}\nonumber\\
    &+&\hspace{-1mm}\Big{|}\Big{|}\disp(\timestamp_i)\hspace{-1mm}-\hspace{-1mm}\disp_{GT}(\timestamp_i)\Big{|}\Big{|}^2\nonumber\\ &+&\lambda_{3D}\Big{|}\Big{|}M(\timestamp_i)-M_{GT}(\timestamp_i)\Big{|}\Big{|}^2 \bigg{)} \hspace{10mm}
\end{eqnarray}

with $||.||$ the L2-norm, $\theta_{GT},\disp_{GT}$ the ground truth body model parameters and $\lambda_{3D}$ a weighting coefficient that controls the relative influence of the per vertex distance. 

The second term $\mathcal{L}_{segment}$ acts as a per segment reconstruction loss, which guarantees that each segment represents a realistic motion and allows for realistic reconstructions where segments are overlapping. 
\small{
\begin{eqnarray}
    \mathcal{L}_{segment}=\frac{1}{mn}\sum_{i=1}^n\sum_{j=1}^m G_j(\timestamp_i)\Big{|}\Big{|} \rigidtransfo_j *\theta_j(\timestamp_i)-\theta_{GT}(\timestamp_i)\Big{|}\Big{|}^2\nonumber\\
    +\frac{1}{mn}\sum_{i=1}^n\sum_{j=1}^m G_j(\disp_i)\Big{|}\Big{|} \rigidtransfo_j *\disp_j(\timestamp_i)-\disp_{GT}(\timestamp_i)\Big{|}\Big{|}^2
\end{eqnarray}
}
\
\section{Evaluation of the motion prior}

We start by outlining the implementation and data used to build our model. To evaluate the motion prior, we test the generalization to sequences of duration outside of the training set, evaluate the influence of the sequential latent representation, and the segmentation learning. Finally, we leverage our prior on a spatio-temporal completion task to evaluate its benefits against state of the art motion priors.
More qualitative results are provided in supplementary material.

\subsection{Implementation and data}
\label{sec:implem}
\textbf{Implementation details}
Our method is implemented using pytorch and the Adam optimizer is used for optimization. SMPL represents a static body by 22 skeleton joints and 16 body shape parameters. We discard the foot joints which have constant rotation in AMASS, which results in 20 skeleton joints. Each joint is represented in 6D using its relative rotation to its parent joint~\cite{zhou2019continuity}. For body shape, we use the first 8 shape components. While our method is flexible~\wrt the number of input frames $n$, we fix $n=100$ for training and train on motions of $3-5s$ by randomly sampling subsequences from the training set. During training, global displacements are normalized to $[-1,1]$ for each direction and timestamps are scaled to $[0,1]$. Unless stated otherwise, we set $m=8$ and each latent vector has dimension $\latentdim=256$. When training the motion prior, we initially use a learning rate of 1e-4, which is reduced to 1e-5 after 20 epochs without improvement of the training loss, and further decreased to 1e-6 after 20 more epochs without improvement. This is done using the plateau scheduler of pytorch. During the training phase we use $\lambda_{KL}=0.0001$,$\lambda_{reg}=0.01$ and set $\lambda_{3D}=0$ for the first 500 epochs because the 3D term slows down training significantly. Once we obtain good convergence after 500 epochs, we set $\lambda_{3D}=1$ for 500 epochs. This significantly increases the pressure on trajectory reconstruction and gives a hierarchical importance to the joints, greatly reducing the reconstruction error in $mm$. We use a batch size of 16. The training phase takes between 1 and 2 days on a  Geforce RTX 2080Ti with 12GB RAM.

\textbf{Data splits}
AMASS~\cite{mahmood2019amass} is a collection of different datasets parameterized by SMPL that contains a variety of motions and body shapes. We leave some collections of this dataset for validation ensuring that all experiment are evaluated on unseen motions and unseen body shapes. 


\subsection{Generalization}

\begin{figure}
\includegraphics[width=\columnwidth]{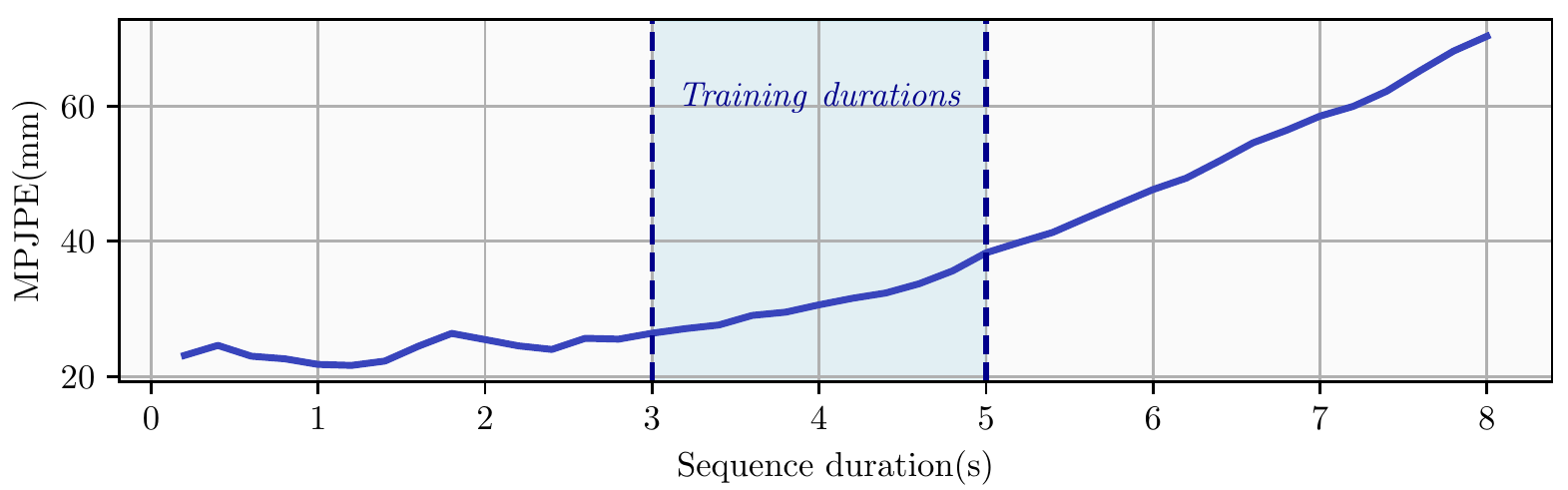}
\caption{Generalization to sequence duration outside training set (for training, duration of $3-5s$). Blue line shows evolution of MPJPE (lower is better) of our method for different sequence duration.}
\label{fig:recvssota}
\end{figure}

Our model is learned using sequences with duration $3-5s$ and $m=8$ latent primitives. We analyze how well this model generalizes to sequences of different duration by applying it to duration $0.2-8s$. To process sequences longer than $5s$ with our method, they are virtually accelerated by scaling the timestamps to $[0,1]$. We perform a forward pass on test sequences of different duration and consider the mean per joint position error (MPJPE) between input and output joints, which is a standard metric introduced in \cite{kocabas2020vibe}. This error is averaged over the sequence. Figure~\ref{fig:recvssota} shows the evolution of MPJPE for different sequence duration. Our model generalizes well to sequences of shorter duration than those used during training, and the error degrades gracefully for sequences of longer duration. 
%

\subsection{Sequential latent representation}
To evaluate the value of learning a sequence of latent primitives instead of a single latent vector, we compare our method to a baseline result when setting $m=1$. To compare the same total number of latent dimensions, we consider two models in this experiment: one trained with $m=4$ where each segment has $\latentdim=256$ latent dimensions and one with $m=1$ and $\latentdim=1024$. For evaluation, we consider the same generalization plot to sequences of different duration as before. Note that $m$ is reduced in this setting for practical reasons as training for $m=1$ with high latent dimension is costly. Figure~\ref{fig:ablsegm} shows that $m=4,\latentdim=256$ leads to significantly lower MPJPE than $m=1,\latentdim=1024$ and generalizes significantly better to sequences of different duration. This shows that a sequential representation better generalizes to varying motions and duration.

\subsection{Segmentation learning}
\begin{figure}
\includegraphics[width=\columnwidth]{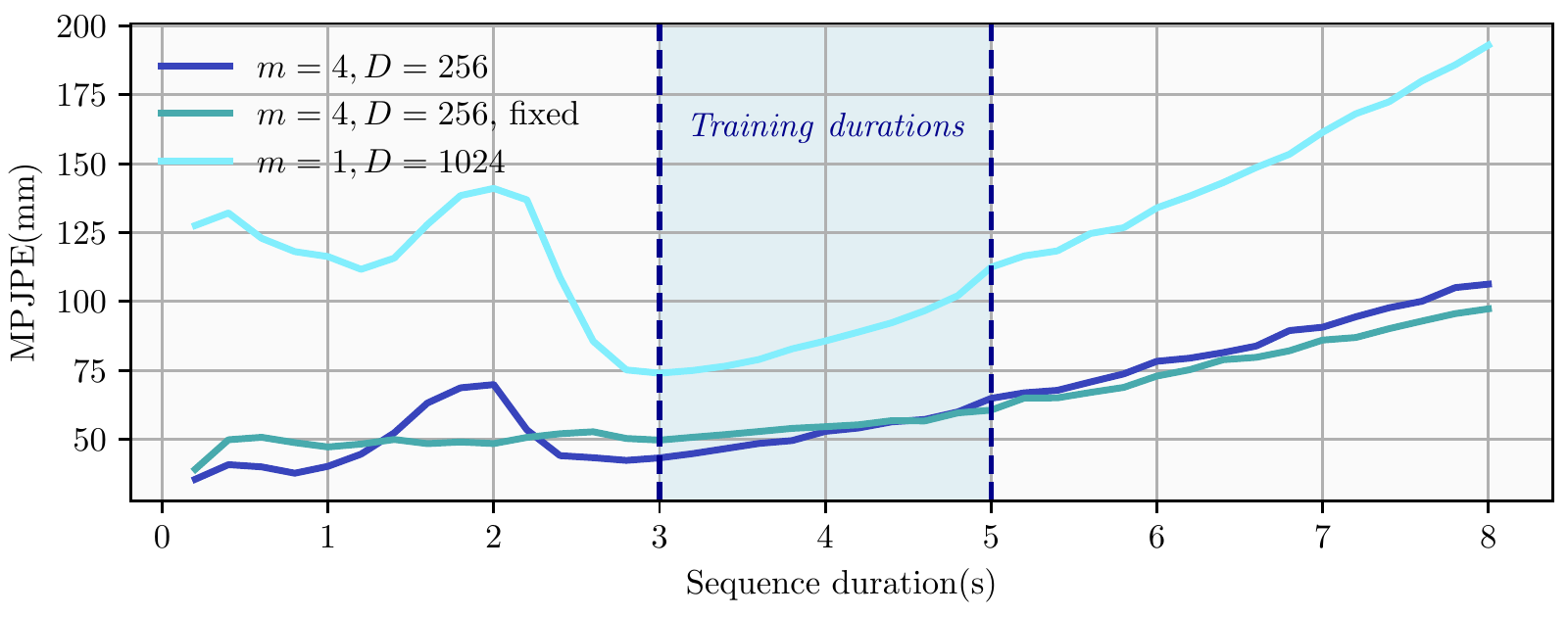}
\caption{Value of using sequences of latent primitives and flexible segmentation. Lines show evolution of MPJPE (lower is better) of our method for different sequence duration ($3-5s$ sequences used for training). Note that using latent sequences with flexible segmentation $m=4,D=256$ performs best in training interval and generalizes well.}
\label{fig:ablsegm}
\end{figure}

To evaluate the influence of learning segment durations, we compare our method to a baseline trained with fixed segmentation parameters $\duration_i=1/m$. As in the previous experiment, we set $m=4,D=256$ for both models and consider the generalization plot to sequences of different duration. Figure~\ref{fig:ablsegm} shows that the differences between the two models are minor. Allowing for flexible segments slightly degrades performance for longer sequences; the virtual acceleration of sequences is more detrimental to the model when learning the segmentation parameters as they are more heavily influenced by timestamp variations. However, flexible segments slightly improve the model performance in the training interval.

\subsection{Comparative evaluation}

We now provide a comparative evaluation on the task of spatio-temporal motion completion \wrt a strong baseline based on a static parametric human body model and two recent state of the art motion priors~\cite{xu2021exploring,li2021task}. A common task to evaluate that motion prior learnt a space of plausible human motion is to perform data completion from sparse 4D inputs by optimizing a latent representation. As most existing methods cannot process sequences of arbitrary duration, our evaluation focuses on how well the methods perform when the input signals are degraded spatially and/or temporally for a given duration. In this experiment, we consider unordered pointcloud inputs at increasingly sparse spatial and temporal resolutions. 

\textbf{Evaluation protocol}
To prevent any biases from having learned on AMASS, all comparative evaluations are performed on a \emph{multi-view test set}, which was acquired by a multi-view camera system at $50$ frames per second and for which no parametric representation is available. The dense per-frame pointclouds of roughly $10,000$ points were obtained by a multi-view stereo method~\cite{leroy2018shape}. The data consists of 4 subjects (2 males, 2 females) performing different motions including boxing, kicking, sidestepping and various types of walking and running and one cartwheel sequence. In total, $170$ motion sequences are used for testing.
To evaluate the robustness of the methods~\wrt degraded input signals, we downsample the pointcloud sequences spatially to $100$ and $1000$ points per frame, and temporally to $5$ and $10$ fps. For each experiment, we reconstruct coherent 4D sequences at $30$ fps which is the proposed framerate in \cite{xu2021exploring} and \cite{li2021task} and evaluate the error by the mean Chamfer distance over all frames of all test sequences. As the closest state of the art methods consider sequences of fixed duration ($4s$ and $2s$, respectively), we perform our evaluation of sequences of duration $4s$, optimizing two latent vectors for \cite{li2021task}. All comparisons are based on code and pre-trained models provided with the respective publications. 

\textbf{Parametric baseline VPoser+SLERP}
The first baseline relies on the state of the art static pose prior VPoser~\cite{pavlakos2019expressive}. In this approach, a latent pose representation is output per frame. As the global displacement is not encoded in VPoser, we additionally optimize per frame displacement. To increase the temporal resolution, we linearly interpolate between observed frames for displacement and using spherical linear interpolations (SLERP) for body pose rotations. We call this baseline VPoser+SLERP in the following. 

\textbf{Motion prior with frequency guidance~\cite{xu2021exploring}}
We compare to two parametric human motion priors. The first one uses frequency guidance and was trained for motions of fixed duration ($4s$) at $30$ fps. This prior does not encode global displacements, so we optimize them per input frame and interpolate linearly for the remaining frames.

\begin{figure*}
    \centering
    \includegraphics{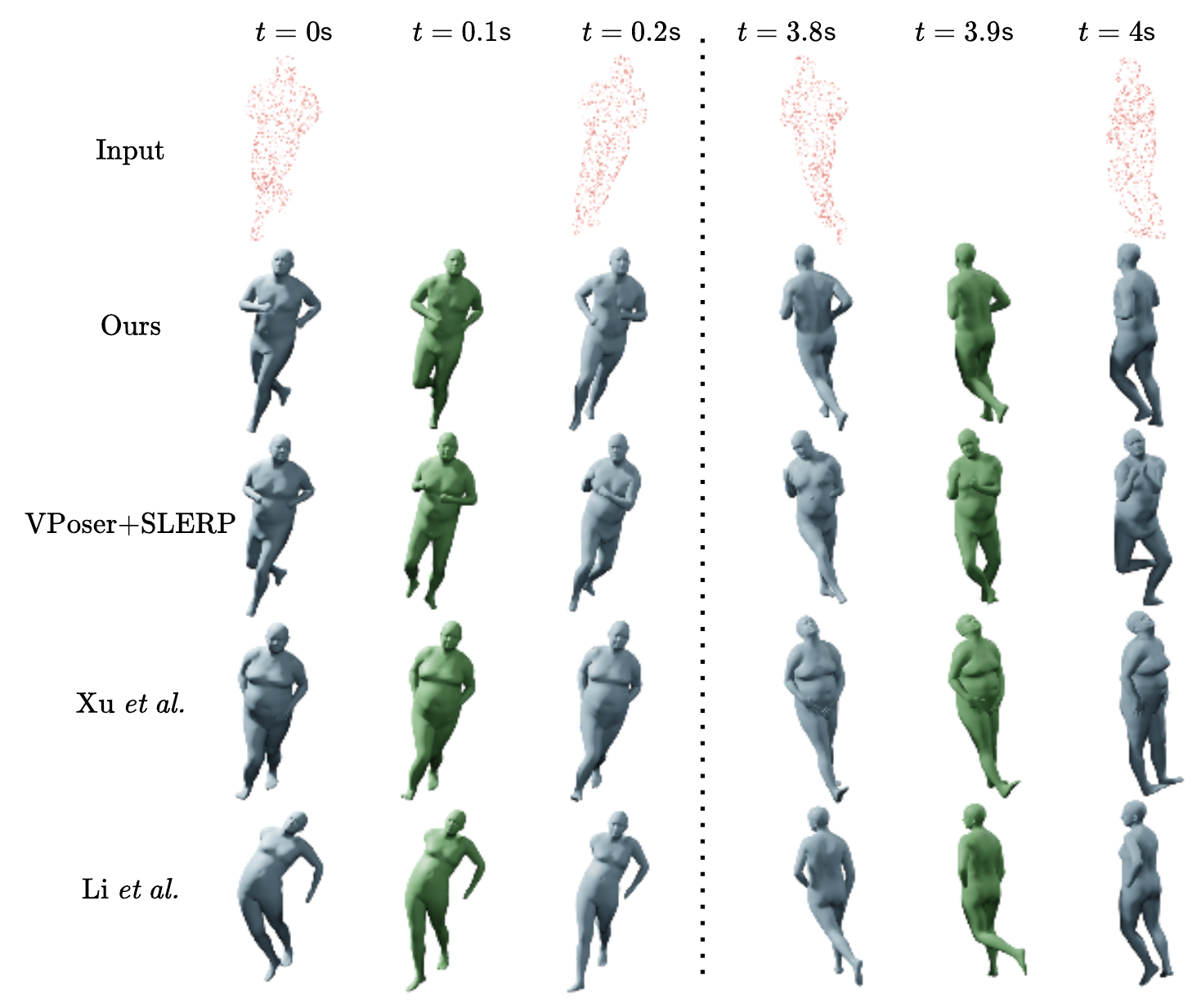}
    \caption{Comparison to state of the art on a challenging example. Completion results on a sequence of a person running in circle; we show frames close to the beginning and the end of the sequence. Our method estimates pose more precisely than other strategies. Blue meshes approximate input frames, green meshes are interpolated by the motion priors.}
    \label{fig:res_completion}
\end{figure*}

\begin{figure*}
    \centering
    \includegraphics[width=0.9\textwidth]{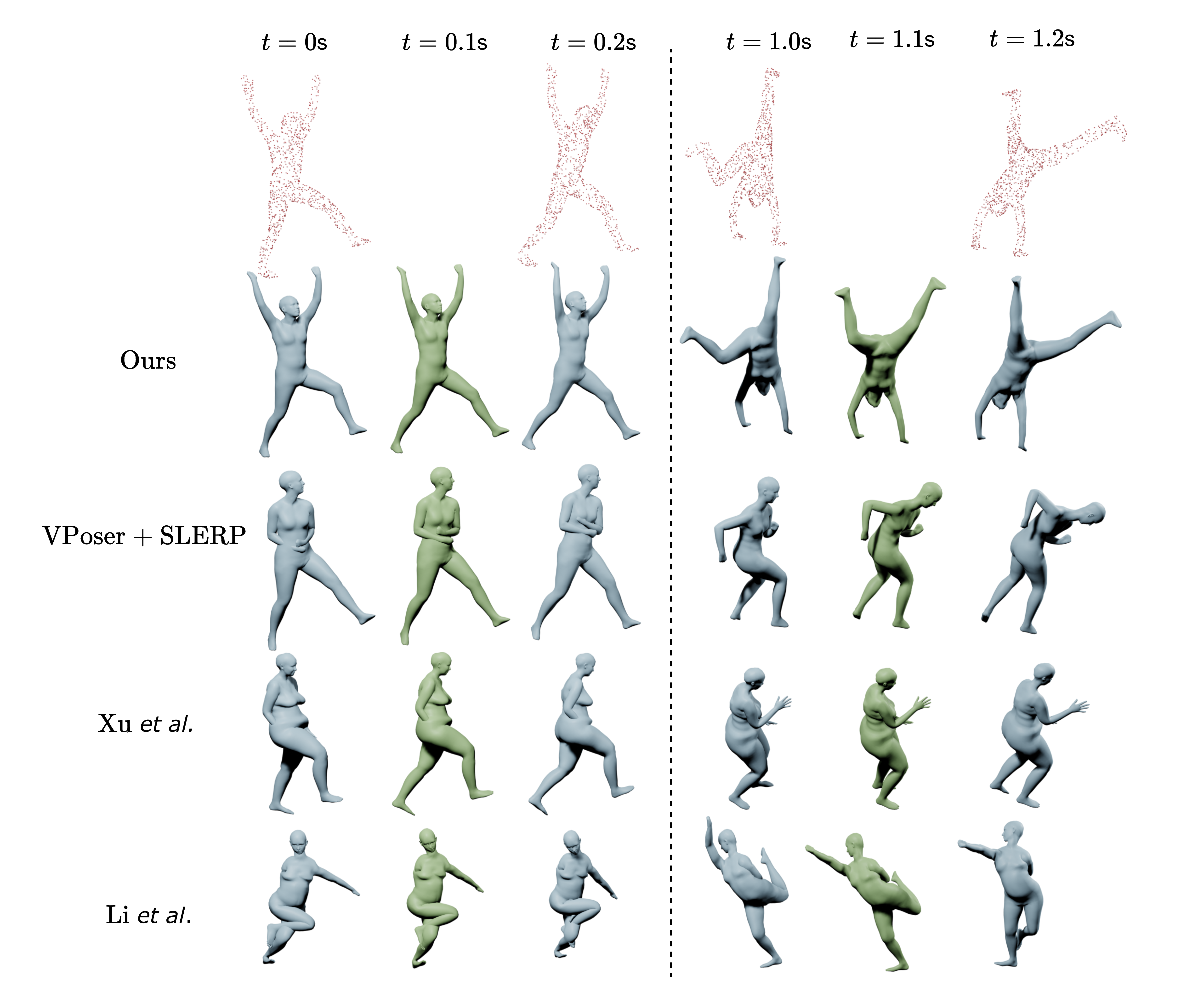}
    \caption{Comparison to state of the art on the most challenging example of the test set: a cartwheel; same color coding as Figure \ref{fig:res_completion}.}
    \label{fig:res_completion2}
\end{figure*}
\textbf{Hierarchical motion prior~\cite{li2021task}}
The second parametric prior uses a hierarchical approach to encode motions of fixed duration ($2s$) at $30$ fps. As we consider sequences of $4s$ in our comparisons, we optimize for two segments by using 
two non-overlapping sliding windows of 2s.

\textbf{Latent optimization}
Given a sequence of sparse pointclouds $\{P_i(\timestamp_i),\timestamp_i\}_{i=1}^n$ we optimize for a latent representation $\{\{z_j\}_{j=1}^m,\beta\}_{i=1}^n$ that best explains the observation. To initialize the latent representation, we randomly sample from the prior distribution in latent space for Xu \etal \cite{xu2021exploring}, Li \etal \cite{li2021task} and VPOSER+SLERP.  

For our method, a sequence characterizing a plausible motion cannot be initialized by independently sampling from the latent prior per latent primitive. Therefore, we use an initialization encoder which is trained as a mapping function from point cloud sequences to the sequential latent space. This initialization encoder is trained using pointcloud representations of the AMASS sequences. The initialization encoder has a similar architecture to the encoder network of our model shown in Figure~\ref{fig:encoder} but the embedding layer has been replaced by a PointNet~\cite{qi2017pointnet}. We use the output of this pointcloud encoder as initialization.

Then we solve for $\underset{\{z_j\}_{j=1}^m,\beta}{argmin}(\mathcal{L}_{comp})$ with: 
\begin{eqnarray}
    \mathcal{L}_{comp}=\mathcal{L}_{chamfer}(M(\mathcal{D}(\{z_j\}_{j=1}^m,\beta,\timestamp_i),P_i(\timestamp_i))\nonumber\\
            + \lambda_{prior}\mathcal{L}_{prior}(\{z_j\}_{j=1}^m),
\end{eqnarray}
where $\mathcal{L}_{chamfer}$ is the average Chamfer distance between input pointclouds and the vertices of the output meshes. $\mathcal{L}_{prior}$ constrains the latent primitives to stay in the latent distribution. For Xu \etal \cite{xu2021exploring}, Li \etal \cite{li2021task} and VPOSER+SLERP, it constrains the latent representation to stay close to the origin. In our method it constrains the latent representation to stay close to its initialization. We set $\lambda_{prior}=0.01$ for all methods.

\begin{table}[b]
    \centering
    \begin{tabular}{|c||c|c||c|c||c|c|}
        \hline
         \# points per frame & \multicolumn{2}{c||}{100} & \multicolumn{2}{c||}{1000} & \multicolumn{2}{c|}{10000}  \\\hline
         Input fps & 5 & 10 & 5 & 10 & 5 & 10\\ 
         \hline
         \hline
         VPoser+SLERP & 27&25&24&20&24&20\\
         \hline
         Xu~\etal~\cite{xu2021exploring}& 28&26&26&24&26&24 \\
         \hline
         Li~\etal~\cite{li2021task}& 83&79&52&48&42&40\\
         \hline
         Ours & \textbf{21}&\textbf{17}&\textbf{17}&\textbf{13}&\textbf{17}&\textbf{13} \\\hline
    \end{tabular}
    \vspace{1mm}
    \caption{Comparison to state of the art using average Chamfer distance ($mm$) (lower is better). Motion completion from different spatial (\# points) and temporal (fps) resolutions.}
    \label{tab:my_label}
\end{table}

\textbf{Results and discussion}
Table~\ref{tab:my_label} reports the results obtained when completing from 5 and 10fps motion to a 30fps motion, when considering input sequences increasingly sparsely sampled in space and time. Note that our method outperforms existing methods by a large margin, especially when considering very sparsely sampled input, and degrades gracefully for decreasing input resolutions. 

Figures~\ref{fig:res_completion} and \ref{fig:res_completion2} show qualitative results on a sequence of a person running in a circle and a challenging cartwheel sequence when comparing the different methods. While our method generates a resulting motion very close to the sparsely sampled input point clouds, VPOSER+SLERP and Xu \etal. fail to capture the global orientation of the motion, because they do not encode global displacement in their latent representation. While Li \etal. finds the correct global orientation for the running sequence, the poses towards the beginning of the sequence are unrealistic. In contrast, our method is robust to global orientation and translation. As it was trained to predict the rigid transformation between segments $\rigidtransfo_i$ and encodes global displacement in latent space, it does not suffer from discontinuities at segment transitions. It also learned detailed motion features thanks to the small temporal windows covered per segment, which is especially visible on the cartwheel motion where our model correctly captured arm positions while other methods could not generalize to this challenging input. 

\section{Conclusion and future works}

This work presented a temporally implicit spatio-temporal representation of motion using a sequence of latent primitives. We showed that using latent primitives characterizing temporal segments of motion allows for a gain in precision which outweighs the gain of adding more latent dimensions. We also showed that our method learned plausible human motion and outperformed state-of-the-art motion priors on a completion task on unstructured pointclouds. 

Future work will investigate sequential learning which takes into account dependencies  between  motion  primitives. It could prove useful  for motion synthesis and be promising to generate a coherent sequence of primitives from scratch.

\section*{Potential negative societal impact}

This work presents a novel representation of human motion sequences using a sequence of latent primitives, and an approach that allows to recover detailed human data from very sparse input. As detailed 3D human motion data is highly personal, such an approach could be used maliciously. Furthermore, our model might be used to retarget motions to different body shapes by changing $\beta$, which could be used to generate disinformation.

\section{Acknowledgements}

We thank Diego Thomas and Raphaël Dang Nhu for interesting discussions about the method design and paper redaction. We also thank Julien Pansiot and Laurence Boissieux for managing the acquisition and data processing of the multi-view test set. 
This work was supported by French government funding managed by the National Research Agency under the Investments for the Future program (PIA) grant ANR-21-ESRE-0030 (CONTINUUM) and 3DMOVE - 19-CE23-0013-01. 


{\small
\bibliographystyle{ieee_fullname}
\bibliography{biblio}

\begin{thebibliography}{10}\itemsep=-1pt

\bibitem{Akhter2012}
Ijaz Akhter, Tomas Simon, Sohaib Khan, Iain Matthews, and Yeaser Skeikh.
\newblock Bilinear spatiotemporal basis models.
\newblock {\em ToG}, 31:\#17:1--12, 2012.

\bibitem{Anguelov05}
Dragomir Anguelov, Praveen Srinivasan, Daphne Koller, Sebastian Thrun, Jim
  Rodgers, and James Davis.
\newblock {SCAPE}: shape completion and animation of people.
\newblock {\em Transactions on Graphics}, 24(3):408--416, 2005.

\bibitem{boukhayma18}
A. Boukhayma and E. Boyer.
\newblock Surface motion capture animation synthesis.
\newblock {\em TVCG}, 2018.

\bibitem{ghorbani2020probabilistic}
Saeed Ghorbani, Calden Wloka, Ali Etemad, Marcus~A Brubaker, and Nikolaus~F
  Troje.
\newblock Probabilistic character motion synthesis using a hierarchical deep
  latent variable model.
\newblock In {\em Computer Graphics Forum}, volume~39, pages 225--239, 2020.

\bibitem{habermann2021}
Marc Habermann, Lingjie Liu, Weipeng Xu, Michael Zollhoefer, Gerard Pons-Moll,
  and Christian Theobalt.
\newblock Real-time deep dynamic characters.
\newblock {\em TOG}, 40(4):94:1--16, 2021.

\bibitem{jiang2021learning}
Boyan Jiang, Yinda Zhang, Xingkui Wei, Xiangyang Xue, and Yanwei Fu.
\newblock Learning compositional representation for 4d captures with neural
  {ODE}.
\newblock In {\em Conference on Computer Vision and Pattern Recognition}, 2021.

\bibitem{jiang2022h4d}
Boyan Jiang, Yinda Zhang, Xingkui Wei, Xiangyang Xue, and Yanwei Fu.
\newblock {H4D:} human 4d modeling by learning neural compositional
  representation.
\newblock In {\em Conference on Computer Vision and Pattern Recognition}, 2022.

\bibitem{kanazawa19}
Angjoo Kanazawa, Jason~Y. Zhang, Panna Felsen, and Jitendra Malik.
\newblock Learning 3d human dynamics from video.
\newblock In {\em CVPR}, 2019.

\bibitem{kocabas2020vibe}
Muhammed Kocabas, Nikos Athanasiou, and Michael~J Black.
\newblock Vibe: Video inference for human body pose and shape estimation.
\newblock In {\em Conference on Computer Vision and Pattern Recognition}, 2020.

\bibitem{leroy2018shape}
Vincent Leroy, Jean-S{\'e}bastien Franco, and Edmond Boyer.
\newblock Shape reconstruction using volume sweeping and learned
  photoconsistency.
\newblock In {\em European Conference on Computer Vision}, pages 781--796,
  2018.

\bibitem{li2021task}
Jiaman Li, Ruben Villegas, Duygu Ceylan, Jimei Yang, Zhengfei Kuang, Hao Li,
  and Yajie Zhao.
\newblock Task-generic hierarchical human motion prior using vaes.
\newblock {\em Conference on 3D Vision}, 2021.

\bibitem{Loper14}
Matthew Loper, Naureen Mahmood, and Michael Black.
\newblock {MoSh:} motion and shape capture from sparse markers.
\newblock {\em ToG}, 33, 2014.

\bibitem{loper2015smpl}
Matthew Loper, Naureen Mahmood, Javier Romero, Gerard Pons-Moll, and Michael~J
  Black.
\newblock {SMPL:} a skinned multi-person linear model.
\newblock {\em Transactions on Graphics}, 34(6):1--16, 2015.

\bibitem{mahmood2019amass}
Naureen Mahmood, Nima Ghorbani, Nikolaus~F Troje, Gerard Pons-Moll, and
  Michael~J Black.
\newblock Amass: Archive of motion capture as surface shapes.
\newblock In {\em International Conference on Computer Vision}, 2019.

\bibitem{3DV118}
Mathieu Marsot, Stefanie Wuhrer, Jean-Sebastien Franco, and Stephane Durocher.
\newblock A structured latent space for human body motion generation, 2021.

\bibitem{niemeyer2019occupancy}
Michael Niemeyer, Lars Mescheder, Michael Oechsle, and Andreas Geiger.
\newblock Occupancy flow: 4d reconstruction by learning particle dynamics.
\newblock In {\em International Conference on Computer Vision}, 2019.

\bibitem{pavlakos2019expressive}
Georgios Pavlakos, Vasileios Choutas, Nima Ghorbani, Timo Bolkart, Ahmed~AA
  Osman, Dimitrios Tzionas, and Michael~J Black.
\newblock Expressive body capture: 3d hands, face, and body from a single
  image.
\newblock In {\em Conference on Computer Vision and Pattern Recognition}, 2019.

\bibitem{petrovich2021action}
Mathis Petrovich, Michael~J Black, and G{\"u}l Varol.
\newblock Action-conditioned 3d human motion synthesis with transformer {VAE}.
\newblock In {\em International Conference on Computer Vision}, 2021.

\bibitem{qi2017pointnet}
Charles~R Qi, Hao Su, Kaichun Mo, and Leonidas~J Guibas.
\newblock {PointNet:} deep learning on point sets for 3d classification and
  segmentation.
\newblock In {\em Conference on Computer Vision and Pattern Recognition}, 2017.

\bibitem{regateiro19}
J. Regateiro, A. Hilton, and M. Volino.
\newblock Dynamic surface animation using generative networks.
\newblock In {\em 3DV}, 2019.

\bibitem{regateiro2021deep4d}
Jo{\~a}o Regateiro, Marco Volino, and Adrian Hilton.
\newblock Deep4d: A compact generative representation for volumetric video.
\newblock {\em Front. Virtual Reality}, 2:739010, 2021.

\bibitem{rempe2021humor}
Davis Rempe, Tolga Birdal, Aaron Hertzmann, Jimei Yang, Srinath Sridhar, and
  Leonidas~J Guibas.
\newblock Humor: 3d human motion model for robust pose estimation.
\newblock In {\em International Conference on Computer Vision}, 2021.

\bibitem{rempe2020caspr}
Davis Rempe, Tolga Birdal, Yongheng Zhao, Zan Gojcic, Srinath Sridhar, and
  Leonidas~J Guibas.
\newblock {CASPR:} learning canonical spatiotemporal point cloud
  representations.
\newblock {\em Advances in neural information processing systems},
  33:13688--13701, 2020.

\bibitem{vaswani2017attention}
Ashish Vaswani, Noam Shazeer, Niki Parmar, Jakob Uszkoreit, Llion Jones,
  Aidan~N Gomez, {\L}ukasz Kaiser, and Illia Polosukhin.
\newblock Attention is all you need.
\newblock {\em Advances in neural information processing systems}, 30, 2017.

\bibitem{xu2021exploring}
Jiachen Xu, Min Wang, Jingyu Gong, Wentao Liu, Chen Qian, Yuan Xie, and
  Lizhuang Ma.
\newblock Exploring versatile prior for human motion via motion frequency
  guidance.
\newblock In {\em Conference on 3D Vision}, 2021.

\bibitem{zhang19}
Jason~Y. Zhang, Panna Felsen, Angjoo Kanazawa, and Jitendra Malik.
\newblock Predicting 3d human dynamics from video.
\newblock In {\em ICCV}, 2019.

\bibitem{Zhang:ICCV:2021}
Siwei Zhang, Yan Zhang, Federica Bogo, Pollefeys Marc, and Siyu Tang.
\newblock Learning motion priors for 4d human body capture in 3d scenes.
\newblock In {\em International Conference on Computer Vision (ICCV)}, Oct.
  2021.

\bibitem{zhou2019continuity}
Yi Zhou, Connelly Barnes, Jingwan Lu, Jimei Yang, and Hao Li.
\newblock On the continuity of rotation representations in neural networks.
\newblock In {\em Conference on Computer Vision and Pattern Recognition}, 2019.

\end{thebibliography}
}

\end{document}